\documentclass{article}
\usepackage{spconf,amsmath,graphicx}
\usepackage{xcolor}

\title{Consistent Mesh Colors for Multi-View Reconstructed 3D Scenes}
\name{ Mohamed Dahy Elkhouly$^{\star,\ddagger}$,
        Alessio {Del Bue}$^{\star,\dagger}$, Stuart James$^{\star}$\thanks{This project has received funding from the European Union's Horizon 2020 research and innovation programme under grant agreement No 870743.}}
\address{$^{\star}$Visual Geometry and Modelling (VGM) Lab, Istituto Italiano di Tecnologia (IIT), Italy \\
$^{\dagger}$Pattern Analysis and Computer Vision (PAVIS), Istituto Italiano di Tecnologia (IIT), Italy \\
$^{\ddagger}$Universit\`{a} degli studi di Genova, Italy}
\begin{document}
%
\maketitle
\begin{abstract}
We address the issue of creating consistent mesh texture maps captured from scenes without color calibration. We find that the method for aggregation of the multiple views is crucial for creating spatially consistent meshes without the need to explicitly optimize for spatial consistency. We compute a color prior from the cross-correlation of observable view faces and the faces per view to identify an optimal per-face color. We then use this color in a re-weighting ratio for the best-view texture, which is identified by prior mesh texturing work, to create a spatial consistent texture map. Despite our method not explicitly handling spatial consistency, our results show qualitatively more consistent results than other state-of-the-art techniques while being computationally more efficient. We evaluate on prior datasets and additionally Matterport3D showing qualitative improvements.
\end{abstract}
\section{Introduction}
\begin{figure}[t!]
   \includegraphics[width=\linewidth]{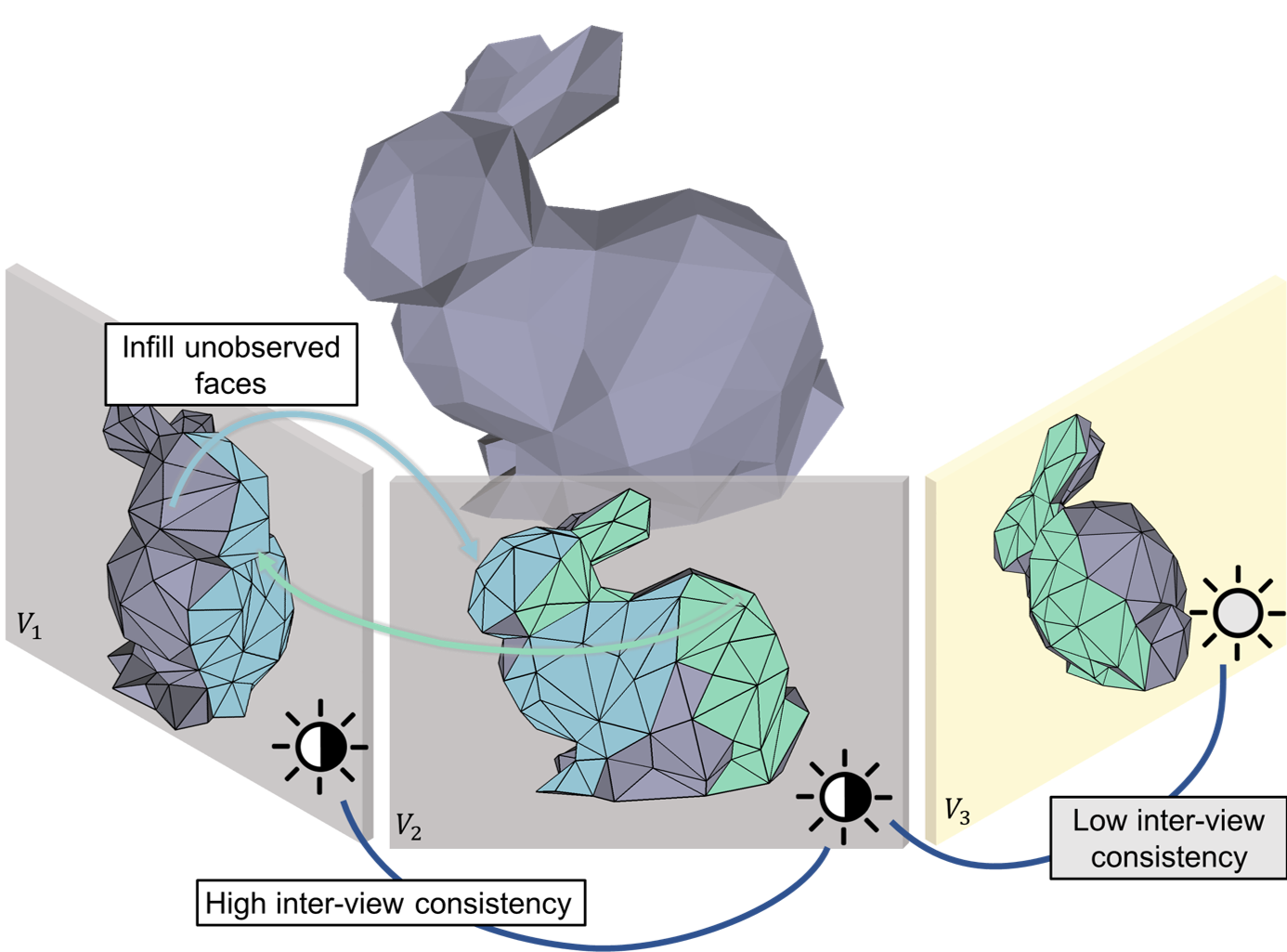}
   \caption{
   We compute a inter-view consistency between jointly observed faces in images, and use the color agreement as a weighting term, this is used to infill across views and weight their contribution. The final per-face color is the trimmed mean across all views to remove remaining outliers.}
    \label{fig:concept}
\end{figure}
 
In recent years there have been many techniques for improving the geometric accuracy of 3D meshes constructed from images, while approaches for the visual appearance have largely been left to traditional averaging or texture mapping approaches. The optimal mesh color or texture will be firstly accurate -- a challenge in an uncalibrated environments; and secondly consistent -- without disjoint patterns occurring across the faces. Such objectives are difficult as during capture lighting and other environmental factors are difficult to control which frequently create artifacts in the reconstruction. However, if such a mesh was to be created it is highly beneficial to downstream task such as relighting.

Most approaches focus on the optimization of the texture \cite{Gal2010CGF, Waechter2014springer, Pages2014CGF} to hide inconsistencies or seams which occur between faces. A seam occurs when the selected texture or color on one face, has a different distribution to the adjacent face. To overcome these issues we rely on the ``inter'' and ``intra'' -view relation of the colors as a cross-correlation factor (see fig.~\ref{fig:concept}). The refined face color can then be used to optimize a texture map creating a consistent texturing across the surface. Neglecting such a relationship results in inconsistent colors from multi-view even using averaging, or using prior mesh texturing methods (see fig.~\ref{fig:results} for examples).
Impressively, unlike the aforementioned methods, which optimize between faces our method has no explicit spatial consistency term, however, the results are spatially stable. Therefore, by utilizing our method as a post-processing step we are able add additional spatial stability to the State-of-the-Art methods.

In this paper, we present a relatively simple technique to provide consistent texturing over a mesh. 
We achieve this by weighting face-based colors extracted from their respective images.
We can list this paper contributions as: (1) Pairwise consistency term based on measuring the cross view face color distribution agreement; (2) Pairwise infilling based re-weighting scheme to implicitly compute the reliability of views; (3) Refinement method for prior mesh textures by correcting the textures using estimated per-face colors.


\section{Related work}
We review the related work in 3D reconstruction for both face (or vertex)-based color as well as texture mapping approaches, which frequently include a view re-weighting scheme. The major body of work is within texture mapping as it provides a richer qualitative result, therefore we include a adaption within our method to apply to texture-mapping as well.

{\noindent\textbf{Face/vertex -based Color:}} The most common approach to face-color estimation, and most computationally efficient, is to use average pixel color across multiple views ~\cite{niessner2013real,Shan20133DV}. However, single pixel values (or surfel -- surface element) are highly sensitive to reconstruction accuracy, which can lead to ghosting or blurring artifacts due to misalignment errors.  Average surfel approaches also don't reject outliers due to environment or illumination values leading to tearing or discontinuity (we show examples of averaging in fig.~\ref{fig:results}). To mitigate this Shan et. al. \cite{Shan20133DV} required images to be captured on a cloud day to help mitigate for extreme intensity values (for shadow or sun). Alternatively, Zhou et al.\cite{Zhou2014ACM} proposed to integrate the color map optimization with the camera pose estimation during reconstruction, by using a Gauss-Newton optimization. However, the approach is computationally expensive and not practical for large scale reconstruction. In contrast, our approach mitigates outlier views through computing the agreement between views, and is computationally efficient in contrast to \cite{Zhou2014ACM}.

{\noindent\textbf{Texture-based:}} As in the case of face-based techniques using a mean textures computed across all views is an intuitive and early technique ~\cite{Frahm2010Eccv}, however, using such an approach results in tearing between the polygons of the mesh.  This problem has been tackled is explicitly tackled in \cite{Gal2010CGF, Waechter2014springer, Pages2014CGF} to optimize the texture selection to avoid seams and color discontinuities. One approach to this exploited Markov Random Fields to resolve this ~\cite{Lempitsky2007}, using an explicit seam visibility calculated by backprojection. This was built on by Waechter et al.~\cite{Waechter2014springer}, which blended the edges of the regions using Poisson editing to smooth out discontinuities and seams, however were still sensitive to the underlying mesh errors.

An alternative to averaging or blending methods, is to select the optimal view of the face as the texture \cite{bi2017patch,rouhani2018multi,Fu_2018_CVPR}. However, the selection criteria is problematic and requires terms for spatial continuity to avoid seams. Wang et. al.~\cite{Wang2018eccv} criteria selected the best-view to provide high detailed texture with a smoothing between textures, however still suffered from inconsistent colors for textures that caused these seams. Best-view based approaches are highly sensitive to illumination variation as this is a complex term to include.

In contrast, our method is able to optimize the output of such texture-based methods, resolving for the color inconsistencies often caused by illumination effects on the texture.
The approach is automatic and does not rely on artist interaction unlike the tools Ptex~\cite{ptex2008} and Mesh colors~\cite{MeshColorsYuksel2010}.

\section{Method}
Our method refines the distribution of color for a given face on a 3D mesh. As input our model takes a set of images $I=\left\{I_1,I_2,I_3,\dots .,I_n\right\}$ where $I_{i}\in {\rm I\!R}^{w\times h \times c}$, the corresponding camera $V=\left\{V_1,V_2,V_3,\dots .,V_n\right\}$ where $V_{i}\in {\rm I\!R}^{3\times 3}$ and a 3D Mesh of faces $f=\left\{f_{1},f_{2},f_{3},\dots .,f_{r}\right\}$. From these inputs we estimate a refined value for the face color $C_{f_i}$ and $T_{f_i}$. We initially compute a pairwise weighting factor using the agreement between views for the jointly observed subset of faces computing a sparse weight matrix ($W$). We then use $W$ to guide the infilling across between views of the set of partially observed faces creating a sparse matrix. To compute $C_{f_i}$ we use trimmed mean over the row to remove remaining outlier values (sec.~\ref{subsec:per_face_colors}). Finally to adjust textures we shift the distribution of $T_{f_i}$ by the ratio between the inferred $C_{f_i}$ and the mean of the texture (sec.~\ref{subsec:correcting_texture_colors}).


\subsection{Face-based color estimation}
\label{subsec:per_face_colors}
For mathematical simplicity, we assume, for the moment, that our images are gray-scale in the case of color images the steps will be calculated for each channel independently.
To calculate the pairwise consistency we compute the union of the observed faces between two frames. Initially, for each view we compute a subset of $f$ for the observed view by re-projecting the face into the camera ($T_{k}=V_{i}^{-1}f_{k}$) we split $f$ into observed $f_{V}$ and unobserved $f_{N}$ from view culling or occlusion using the projection output $T$.  

Therefore, given pair of images $\left(I_i,I_j\right)$ and their corresponding pairs of observations $\left (\left \langle f_{Vi},f_{Ni}  \right \rangle,\left \langle f_{Vj},f_{Nj}  \right \rangle  \right  )$ we compute the union $f_{{O}_{ij}}$ as:
\begin{equation} \label{eq:overlapping} 
  f_{{O}}=f_{{V}_i}\cap f_{{V}_j}~~~\forall ~i\neq j,~~~i,j\in n.
\end{equation} 
To compute the pairwise consistency term $W_{ij}$, we then take $f_{O}$ and the corresponding pixel patches and computer the ratio between the trimmed mean of a corresponding $T_{i}$ and $T_{j}$ in $f_{{V}_i}$ and $f_{{V}_i}$. The trimmed mean truncates the extremes of the distribution (similar to high/low pass filter). Firstly $T$ is sorted, as $T$ is a vector of intensities for a given image channel, and the trimmed mean is computed as:
\begin{equation} \label{eq:trimmean}
\mu_{T}(T) =\frac{1}{\left \lfloor R \right \rfloor}  \sum _{\left \lfloor k \right \rfloor+1}^{n-\left \lfloor k \right \rfloor}{T}
\end{equation} 
where $k=n\alpha$ and $n$ is the number of pixels in the triangle, and $\alpha$ is the trimming factor which we set to $\alpha=0.3$. $R$ is the remaining number of observations after trimming $R= n-2k=n(1-2\alpha)$. The inter-view consistency is therefore the ratio of resultant mean values over jointly observed faces:
\begin{equation} \label{eq:cross_view} 
W_{ij}=\frac{1}{l} \sum_{k=0}^{k=l} 
\frac{\mu_{T}\left(I_{i} \left (T\left (f_{Ok}  \right )  \right )  \right)}
{\mu_{T}\left(I_{j} \left (T\left (f_{Ok}  \right )  \right )  \right)}
\end{equation}
where $l$ is the number of jointly observed face. We then normalize the weights into $[0,1]$ range. The cross-correlation weighting $W$ is therefore a sparse matrix, as not all views will have overlapping observations. The matrix represents the consistency between the views with the intuition that views with high agreement are likely under similar environmental conditions and therefore useful for estimating the color.

With the inter-view weighting established, we aim to compute a matrix $C$ between faces and views by infilling values across views that have a strong consistency. Therefore, we fill the matrix with the observed faces $C_{i}=\mu_{T}\left(T_{i}\right)$, and additionally propagate information from the observed faces $f_{Vj}$ to the unobserved in another $f_{Ni}$ as in:
\begin{equation} \label{eq:2} 
{\hat{C}}_{{j}_{ij}}=f_{{V}_j}\left(f_{{N}_i}\right)W_{ij}.
\end{equation}
The process of infilling the matrix acts as a weighting approach to stabilize the method making the assumption that the higher number of a agreeing views will be the more accurate color. To obtain the final value of each face $f$, we use trimmed mean across the rows of the matrix $C$, as:
\begin{equation} \label{eq:3} 
{\hat{f}}_{k}= \mu_{T}(C_{k}).
\end{equation}
\begin{figure}[t!]
   \includegraphics[width=\linewidth]{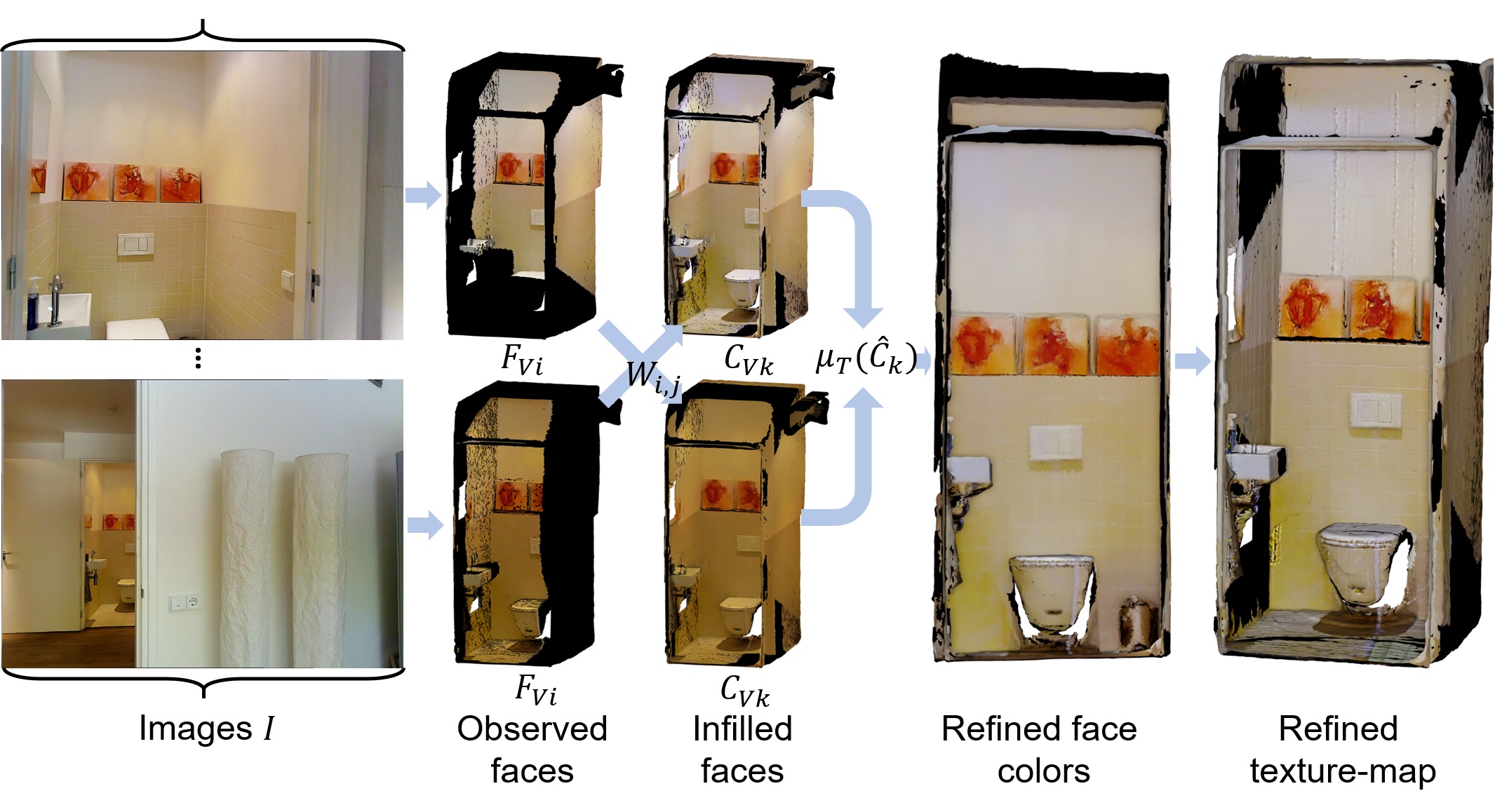}
   \label{fig:ex1}
   \caption{
    From a set of images $I$ and the corresponding 3D mesh as input, we calculate the observed and unobserved faces $\left \langle f_{Vi},f_{Ni}  \right \rangle$ of the mesh. We infill across views $f_{Nj}\mapsto f_{Vi}$ using the weighting $W$. The final face color as the trimmed mean of the infilled faces. We show the result of our infilling, face color estimation and correction of \cite{Waechter2014springer} on a complex part of a scene of the Matterport3D~\cite{Matterport3D} dataset.}
\end{figure}

\subsection{Texture-based mesh correction}
\label{subsec:correcting_texture_colors}
Textures offer a richer visual representation than face-color as they reduce the number of polygons require for qualitatively satisfying result. Therefore, methods focus on optimizing for constancy of texture-mapping  \cite{Waechter2014springer,Fu_2018_CVPR}, we aim to improve on these methods, given the output of our face based method which computes an optimal mean color value for each face. 

Given a texture map as input (and the output of our face-based color method) we compute the corresponding texture patch for each face $u_k$. We then shift the distribution of $u_k$ by the ratio of the mean values of $M_{k}=\mu\left(u_k\right)$ and the face color $\hat{f}_k$ as in:
\begin{equation} \label{eq:new_tex_pix} 
  \hat{u}_{kr}=u_{kr}\frac{\hat{f}_k}{M_{k}},
\end{equation}
where $r$ iterates through the set of colors in the texture patch. While this is a simple method, it is able to stabilize the output of texture-based methods removing for environmental effects which frequently create undesirable artifacts.

\begin{figure*}[t!]
  \centering
  \includegraphics[width=\linewidth]{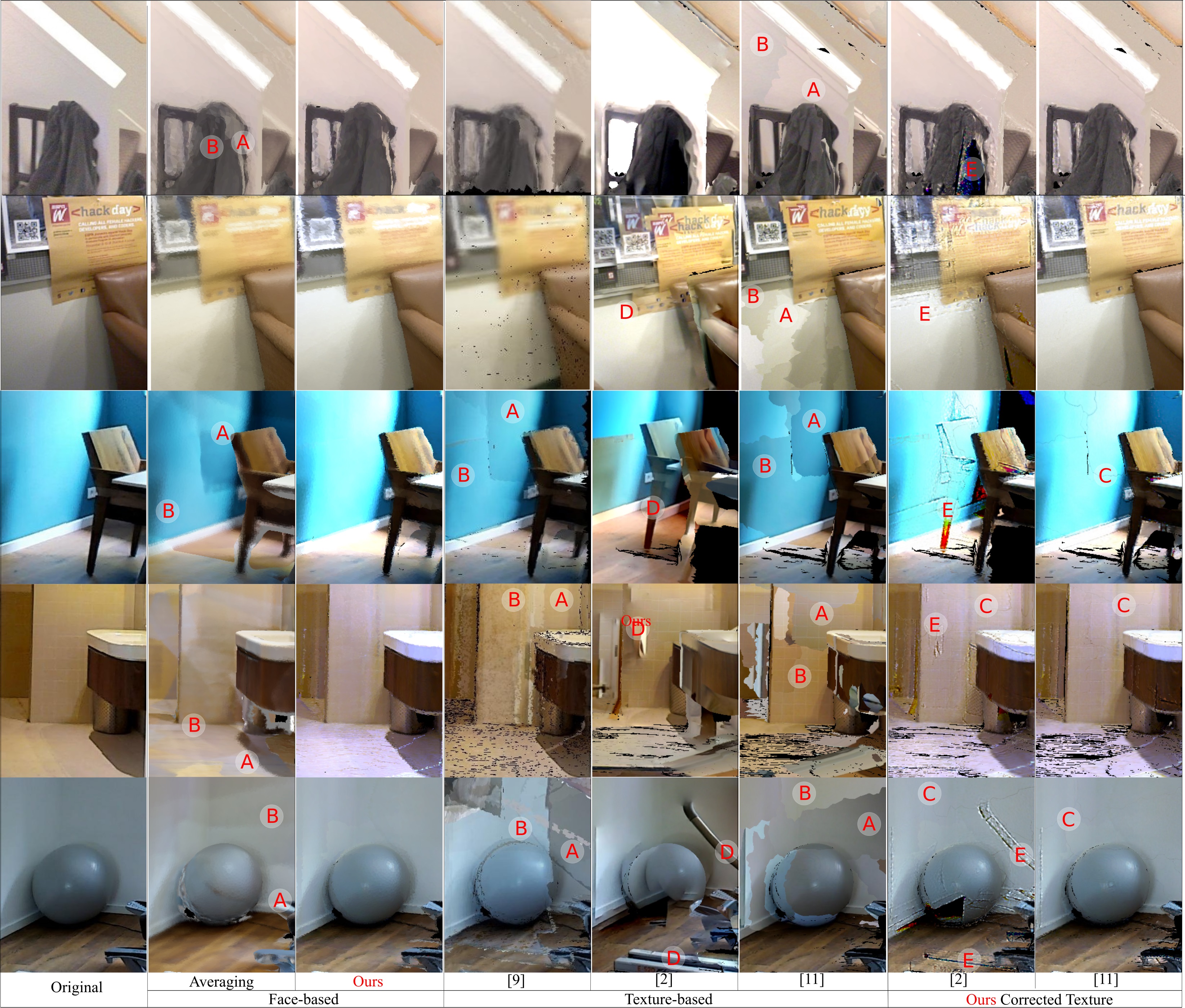}

  \caption{\label{fig:results}
       Comparison between our results as face-based, texture-based and other state-of-art techniques. Scenes from \cite{Fu_2018_CVPR} and \cite{Zhou2013acm} for rows 1 and 2 respectively the remainder are from the  Matterport3D~\cite{Matterport3D}. The marks \textbf{A, B} shows inconsistent color in neighbor regions, \textbf{C} texture seams, \textbf{D} registration issues and the similar  \textbf{E} texture registration issues
      }
\end{figure*}
\section{Evaluation}
We evaluated our method for both face-based colors and texture-based mapping. We show qualitative results in fig.~\ref{fig:results} across three different datasets: \cite{Fu_2018_CVPR} in row one, \cite{Zhou2013acm} in row two, and three examples from Matterport3D~\cite{Matterport3D}. For variation we select scenes from \cite{Fu_2018_CVPR} and \cite{Zhou2013acm} which are of dense (narrow baseline) cameras and are often close to the scene they are capturing with limited scale change. For efficiency we subsample the views, 10\% of original, and decimate the mesh, keeping 20\% of the original. In contrast, Matterport3D offers sparse views with high variation in scale due to a wide baseline. Additional results can be seen in the supplementary material (video and paper). To distinguish visual artifacts we use red letter marks: \textbf{A, B} shows inconsistent color in neighbor regions, \textbf{C} texture seams, \textbf{D} registration issues and the similar  \textbf{E} texture registration issues.


We compare our refined texture mesh against the output from ~\cite{Waechter2014springer,Fu_2018_CVPR}. In addition we show the the original output of \cite{bi2017patch} as we were unable to register the projected image to a texture-map so we could not correct it, however, we include for visual comparison.
Regarding \cite{bi2017patch} we used the available implementation from ~\cite{WU2019oc} and use 10 scales and 50 iterations of alignment per scale. 

From figure~\ref{fig:results} for neighbor consistency (\textbf{A, B}) our refined results (columns 3, 7, 8) show that we successfully overcome this problem in the estimated face-based colors, and in corrected textures. 
We note that Waechter et al. ~\cite{Waechter2014springer} had registration issues (\textbf{C, D}), however, our  algorithm was able to mostly hide this and provide visually closer textures to the original. We note there is inappropriate texture registration (\textbf{E}) in some cases.
Regarding our algorithm we can notice that it sometimes provides inaccurate brightness in contrast to the original image (rows 3,4,5) for brighter.
Moreover, our estimation brightness could be adjusted after aggregation without affecting consistency between colors by adjusting $\alpha$ in the trimmed mean to serve for the application. 
We can also notice that there are some visible seams in our corrected textures (\textbf{C}), adding margins to textures can resolve this as performed by \cite{Waechter2014springer,Fu_2018_CVPR}, however during the correction process we only have the output textures, this cold be resolved by integrating the method or to give access to the margins at output. 
Our method is inherently more suitable to diffuse object, but still our technique provides satisfying results on specular objects as shown on the chair (row 2) and ball (row 5), however this is sensitive to the number of views and the viewing angles.
It is highly likely that increasing the number of faces or number of views, will provide our technique with additional benefit by providing more precise face-based colors which in turn are used for providing extra sharp corrections, however the technique achieves visually satisfying improvement over prior methods.

\section{Conclusions}
We presented an efficient technique to reduce color discontinuities in multi-view 3D meshes by weighting view contribution and adapted it to apply to texture-mapping by shifting the mean of the distribution. We show qualitative improvement on the output textures of State-of-the-Art techniques resolving for registration and environmental issues on prior datasets and scenes from Matterport3D. We do not include explicit spatial consistency, aiding in computational efficiency, however the qualitative results show a spatial consistency demonstrating the stability of the method. 

\bibliographystyle{IEEEbib}
\bibliography{egbib}
\appendix
\begin{figure*}[t!]
  \centering
  \includegraphics[width=1.0\textwidth]{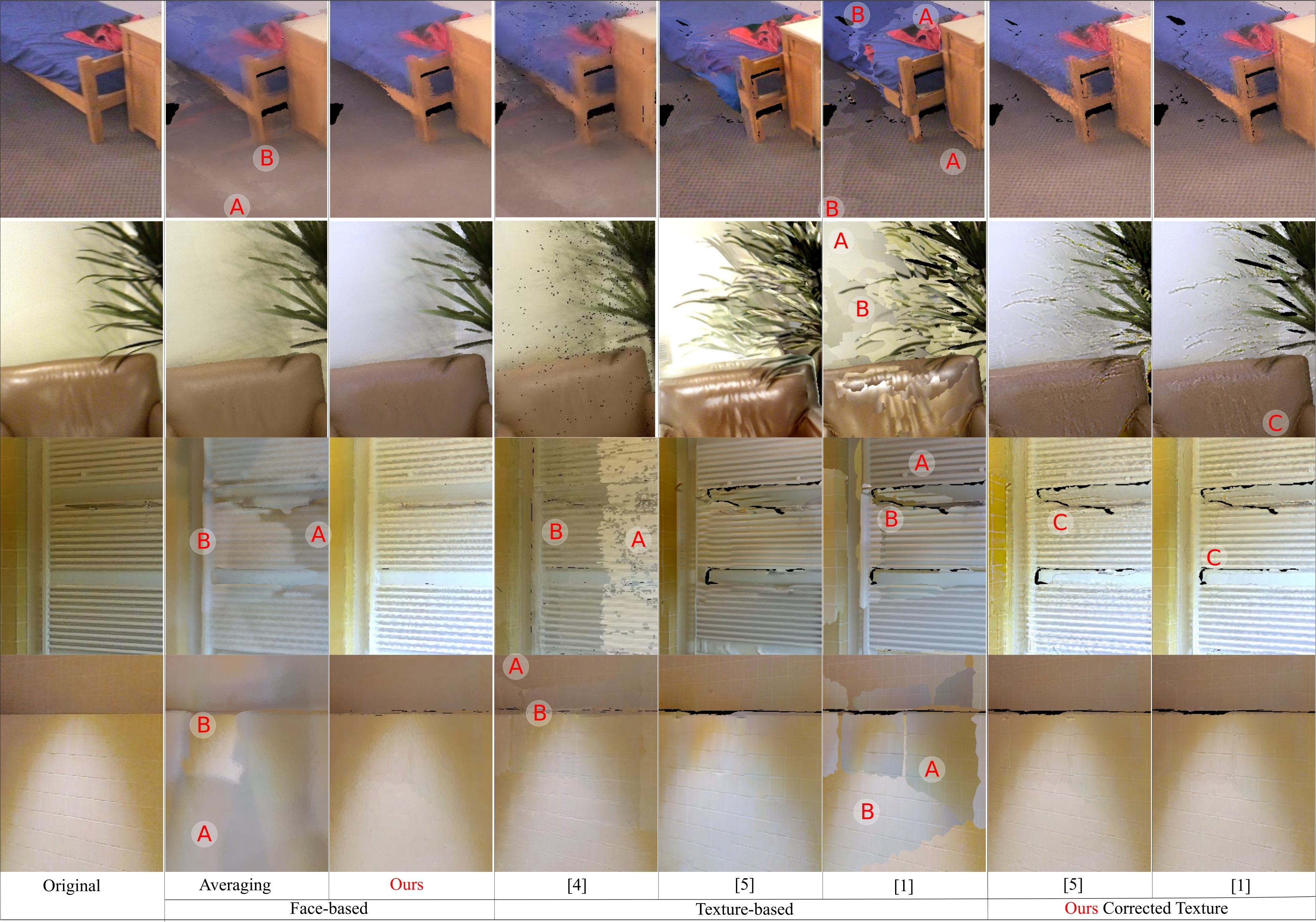}

  \caption{\label{fig:supp_results}
       Comparison between our results as face-based, texture-based and other state-of-art techniques. Scenes from \cite{Fu_2018_CVPR} and \cite{Zhou2013acm} for rows 1 and 2 respectively the remainder are from the  Matterport3D~\cite{Matterport3D}. The marks \textbf{A, B} shows inconsistent color in neighbor regions, \textbf{C} texture seams, \textbf{D} registration issues and the similar  \textbf{E} texture registration issues.
      }
\end{figure*}
\newpage
\section{Supplementary Material}
In addition to the results presented in the main paper, we show another scene of \cite{Fu_2018_CVPR} and \cite{Zhou2013acm} and two more scenes from Matterport3D~\cite{Matterport3D}. We show the results of \cite{bi2017patch,Waechter2014springer,Fu_2018_CVPR} and the refined \cite{Waechter2014springer,Fu_2018_CVPR} of the paper for side-by-side comparison in fig.~\ref{fig:supp_results} (second page). We maintain the same annotation scheme for the paper where marks \textbf{A, B} shows inconsistent color in neighbor regions, \textbf{C} texture seams, \textbf{D} registration issues and the similar  \textbf{E} texture registration issues.

\end{document}